\documentclass[fleqn]{article}

\usepackage{lmodern}
\usepackage[T1]{fontenc}
\usepackage{textcomp}
\usepackage[utf8]{inputenc} 
\usepackage[T1]{fontenc}    
\usepackage{hyperref}       
\usepackage{url}            
\usepackage{booktabs}       
\usepackage{nicefrac}       
\usepackage{microtype}      
\usepackage[utf8]{inputenc} 
\usepackage[T1]{fontenc}    
\usepackage{hyperref}       
\usepackage{url}            
\usepackage{booktabs}       
\usepackage{amsfonts}       
\usepackage{nicefrac}       
\usepackage{microtype}      
\usepackage{amsmath, bm}
\usepackage{systeme}
\usepackage{amssymb}

\usepackage{graphicx}
\usepackage{float}
\usepackage[english]{babel}
\usepackage[utf8]{inputenc}
\usepackage[noend]{algpseudocode}
\usepackage[linesnumbered,ruled,vlined]{algorithm2e}

\usepackage{blindtext}
\usepackage[left=3cm,right=2cm,top=2.5cm,bottom=2cm]{geometry}
\usepackage[table]{xcolor}
\usepackage{tikz}
\usetikzlibrary{positioning}

\makeatletter
\renewcommand\maketitle
  {
   {\Large\bfseries\@title}
   \bigskip\bigskip\bigskip\par\noindent
   {\bfseries\@author}%
   \hfill
   {\large\@date}%
   \bigskip\par\noindent
  }
\makeatother

\newcommand{\naturalto}{%
  \mathrel{\vbox{\offinterlineskip
    \mathsurround=0pt
    \ialign{\hfil##\hfil\cr
      \normalfont\scalebox{1.2}{.}\cr
      $\longrightarrow$\cr}
  }}%
}

\newcommand{\Cross}{\mathbin{\tikz [x=1.4ex,y=1.4ex,line width=.2ex] \draw (0,0) -- (1,1) (0,1) -- (1,0);}}%

\usepackage{xcolor}
\title{Soft Genetic Programming Binary Classifiers}

\author{%
  Ivan Gridin \\
  \texttt{ivan.gridin.pro@gmail.com}
}

\begin{document}

\def\arraystretch{2}%

\maketitle

\bigskip 
\bigskip
\bigskip

\begin{abstract}
The study of the classifier's design and it's usage is one of the most important machine learning areas. With the development of automatic machine learning methods, various approaches are used to build a robust classifier model. Due to some difficult implementation and customization complexity, genetic programming (GP) methods are not often used to construct classifiers. GP classifiers have several limitations and disadvantages. However, the concept of "soft" genetic programming (SGP) has been developed, which allows the logical operator tree to be more flexible and find dependencies in datasets, which gives promising results in most cases. This article discusses a method for constructing binary classifiers using the SGP technique. The test results are \hyperref[sec:results]{presented}. Source code - \url{https://github.com/survexman/sgp_classifier}
\end{abstract}


\bigskip
\bigskip
\bigskip

\section{Introduction}
\label{introduction}

\bigskip

Genetic Programming (GP) is a promising machine learning technique based
on the principles of Darwinian evolution to automatically evolve computer
programs to solve problems. GP is especially suitable for building a classifier of tree representation. GP is a soft computing search technique, which is used to evolve a tree-structured program toward minimizing the fitness value of it. The distinctive features of GP make it very convenient for classification, and the benefit of it is the flexibility, which allows the algorithm to be adapted to the needs of each particular problem.

\bigskip
   
\begin{figure}[H]
\vspace{-2mm}
\centering
    \includegraphics[width = 10cm]{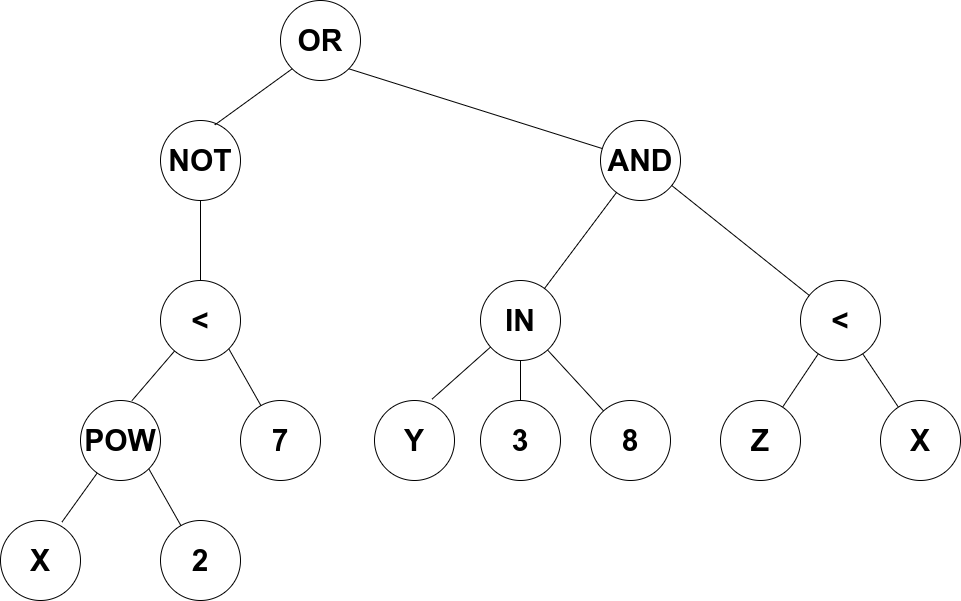}\par 
\caption{Logical tree example}
\label{fig:logical_tree_example}
\vspace{-6mm}
\end{figure}

   A special case of GP studies the logical tree's development as a solution to a classification problem \cite{Kuo2007ApplyingGP}. Logical trees are composed of boolean, comparison, and arithmetic operators, and they output a boolean value ( true or false). A solution presented as a logical tree is a very convenient way to analyze a dataset and interpret a solution. Figure \ref{fig:logical_tree_example} shows an example of logical tree. 

\bigskip

This logical tree can be rewritten as the following system of inequalities:

\[
x^2 \geq 7 \;  \lor \; \systeme{y \geq 3, y \leq 8, z < x}
\]

\bigskip

Evolving logical trees using GP has the benefit of being able to handle both numerical
and categorical data fairly simply. Significant advantage of evolving logical trees is the fact that the trees are highly interpretable to a researcher \cite{Dufourq2015DataCU}. Logical trees are portable and can be easily implemented by various tools and programming languages. Another significant advantage is an ability of feature extraction \cite{GPFeatureSelection2014}.

\medskip

However, one of the most critical problems in logical trees is a crucial logic change when a tree operator changes \cite{math/0612405}\cite{0708.1820}. Such operator changes occur as a result of crossover and mutation operations \cite{Koza}.

\bigskip

\begin{figure}[H]
\vspace{-2mm}
\centering
    \includegraphics[width = 13cm]{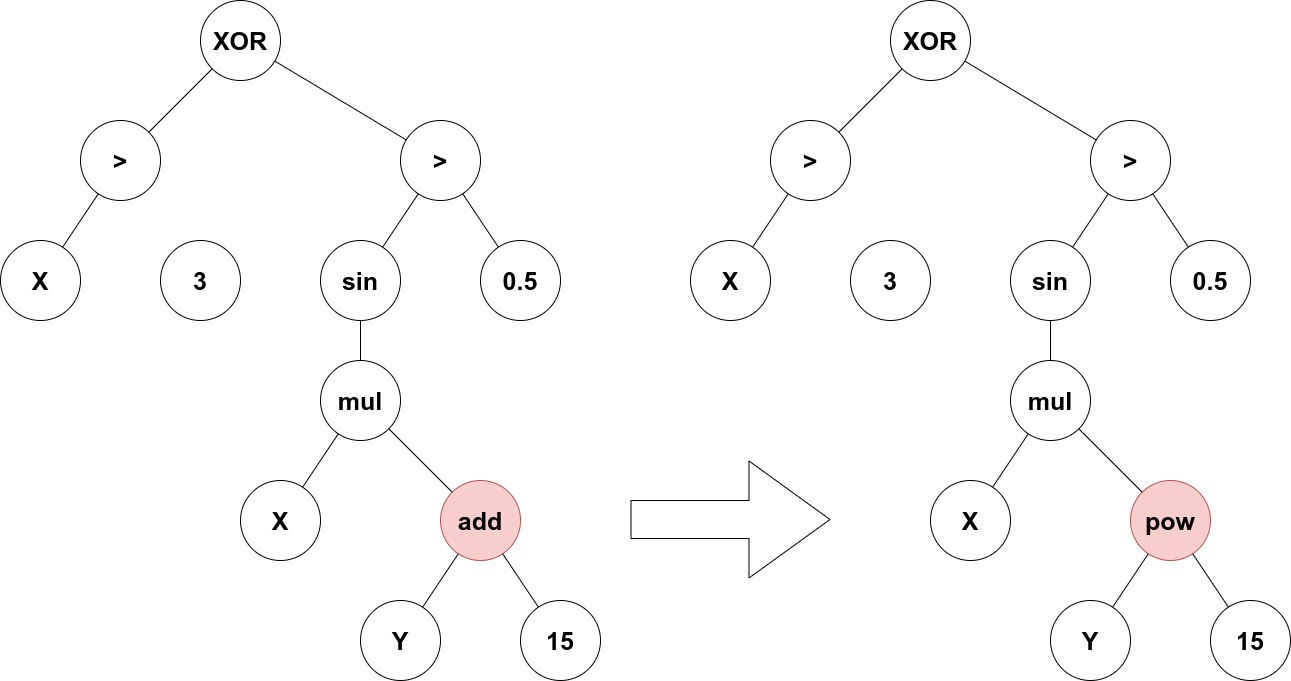}\par 
\caption{Operator change in Logical Tree}
\label{fig:operator_change}
\vspace{-6mm}
\end{figure}

\bigskip
\bigskip

Figure \ref{fig:operator_change} shows a simple change of \textbf{addition} operator to \textbf{power} operator in logical tree:

\[
[x > 3 \;  \oplus \; sin(x(y+15))) > 0.5] \;\; \naturalto \;\; [x > 3 \;  \oplus \; sin(xy^{15}))) > 0.5]
\]

\bigskip
\bigskip

And below at Figure \ref{fig:operator_change_effect} we can see the effect of such modification:

\begin{figure}[H]
\vspace{-2mm}
\centering
    \includegraphics[width = \linewidth]{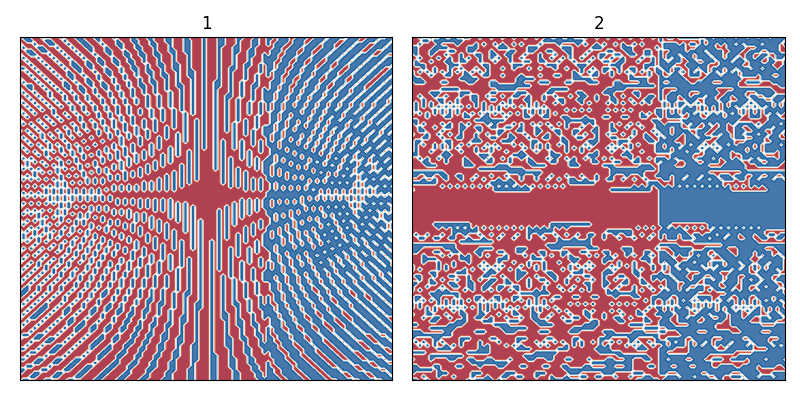}\par 
\caption{Operator change effect}
\label{fig:operator_change_effect}
\vspace{-6mm}
\end{figure}

\bigskip

We see that switching the operator \textbf{+(y,15)} to \textbf{pow(y,15)} changes the logic of the classifier drastically. The more complex the logical tree is, the more this feature is noticeable. This sensitivity of logical trees to tiny changes makes it very difficult to find the appropriate logical tree in a smooth way.

\bigskip
\bigskip
\bigskip
\section{Study Roadmap}

\bigskip

SGP classifier design is based on classical GP classifier design. \hyperref[sec:gp_arch]{First}, we will define the architecture of GP classifier. \hyperref[sec:sgp_arch]{Next}, we will introduce the Soft Genetic Programming approach, its design, and evolution operations. Soft Genetic Programming's main goal is to smoothen evolution improvement and increase the probability of reaching local maxima.  \hyperref[sec:viz]{After}, we will compare the behavior of classical GP Classifier and SGP Classifier. And  \hyperref[sec:results]{finally}, we will present empirical evidence of the applicability of SGP approach.

\clearpage
\newpage

\section{Genetic Programming Classifier Design}
\label{sec:gp_arch}

\bigskip

Genetic programming (GP) is a flexible and powerful evolutionary technique with some special features that are suitable for building a classifier of tree representation. \cite{Koza}.

\bigskip
\subsection{Operators}

We will use the following operators for genetic programming trees \cite{{10.1007/978-3-642-12148-7_1}}:

\bigskip

\textbf{Boolean}:
\begin{align*}
\textrm{OR}:\ \{{0,1}\}^2 \rightarrow \{{0,1}\} \\
\textrm{AND}:\ \{{0,1}\}^2 \rightarrow \{{0,1}\} \\
\textrm{NOT}:\ \{{0,1}\} \rightarrow \{{0,1}\}
\end{align*}

\bigskip

\textbf{Comparison}:
\begin{align*}
\textrm{>}:\ \mathbb{R}^2 \rightarrow \{{0,1}\} \\
\textrm{<}:\ \mathbb{R}^2 \rightarrow \{{0,1}\}
\end{align*}

\bigskip

\textbf{Mathematical}:
\begin{align*}
\textrm{+}:\ \mathbb{R}^2 \rightarrow \mathbb{R} \\
\times:\ \mathbb{R}^2 \rightarrow \mathbb{R} \\
-:\ \mathbb{R} \rightarrow \mathbb{R}
\end{align*}

\bigskip

\textbf{Terms}:
\[
Symbolic:\  \rightarrow X_i \; \; \;   i \in 1,n , \; where \;  n\;  -\;  number\;  of\;  features
\]
\[
Constant:\  \rightarrow \mathbb{R}
\]

\bigskip
\subsection{Random Tree Generation}
\label{subsec:gp_random_tree}

Random tree generation is limited by maximum and minimum operator type subchain \cite{10.1007/3-540-45110-2_69}:

\bigskip

\begin{tabular}{ |c|c|c| } 
 \hline
 & min & max \\
  \hline
 Boolean & 1 & 3 \\
   \hline
 Comparison & 1 & 1 \\
   \hline
 Mathematical & 1 & 4 \\
    \hline
 Terms & 1 & 1 \\
   \hline
\end{tabular}

\bigskip
\bigskip

\subsection{Fitness Function}
\label{subsec:gp_fitness_function}

For fitness function we use \textit{Balanced Accuracy}:

\[
Balanced \; Accuracy = \frac{1}{2}(\frac{TP}{TP + FN} + \frac{TN}{TN + FP})
\]

\bigskip
\subsection{Crossover}

\begin{figure}[H]
\vspace{-2mm}
\centering
    \includegraphics[width = 8cm]{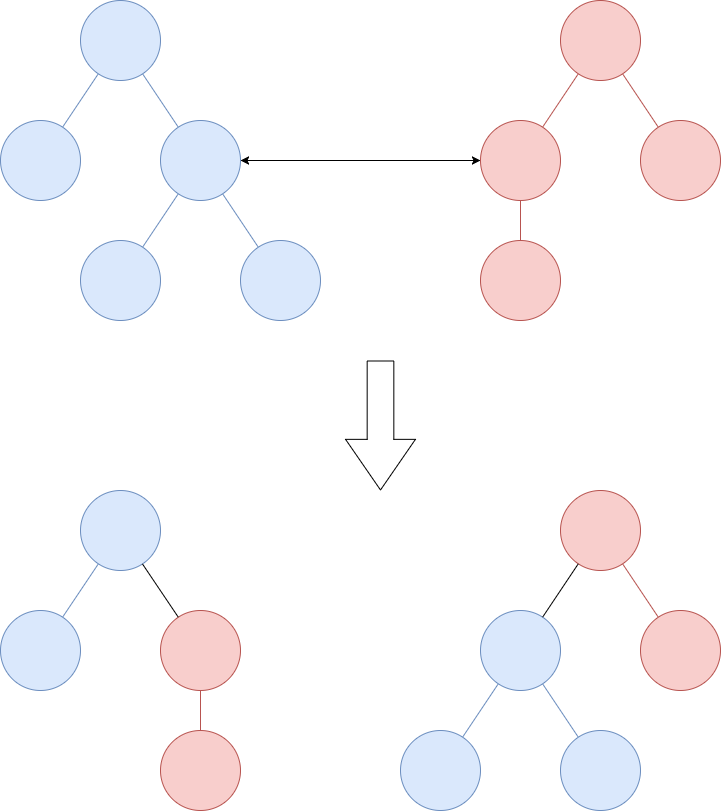}\par 
\caption{Crossover}
\label{fig:crossover}
\vspace{-6mm}
\end{figure}

The crossover operator creates two new offspring which are formed by taking parts
(genetic material) from two parents. The operator selects two parents from
the population based on a selection method. A crossover point is then randomly
selected in both trees, say point $p_1$ and $p_2$ , from tree $t_1$ and $t_2$ respectively. The crossover then happens as follows: the subtree rooted at $p_1$ is removed from $t_1$ and inserted into the position $p_2$ in $t_2$. The same logic applies to the point $p_2$ ; the subtree root at the point is removed from $t_2$ and inserted into the place of $p_1$ in $t_1$ \cite{Dufourq2015DataCU}. Figure \ref{fig:crossover} illustrates the crossover operation. 

\bigskip

\clearpage
\newpage

\subsection{Operator Mutation}

\begin{figure}[H]
\vspace{-2mm}
\centering
    \includegraphics[width = 6cm]{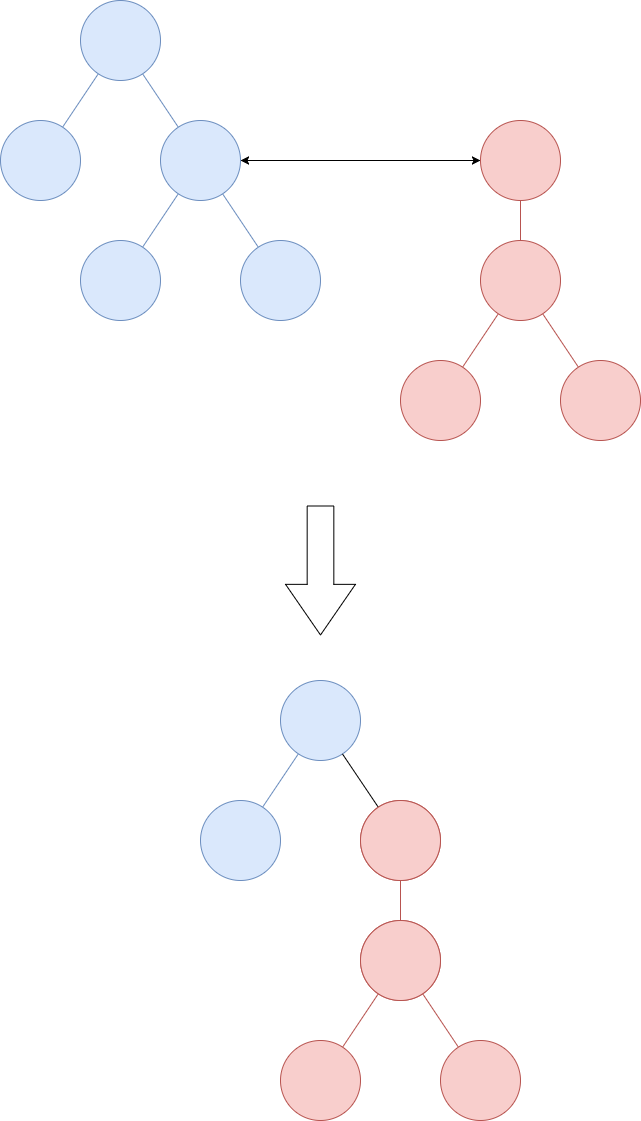}\par 
\caption{Mutation}
\label{fig:mutation}
\vspace{-6mm}
\end{figure}

An Operator Mutation works the following way: an operator is randomly selected and replaced with a random tree. A new randomly generated subtree is inserted instead of selected operator \cite{Koza}. Figure \ref{fig:mutation} illustrates the operator mutation operation.

\bigskip
\bigskip

\subsection{Term Mutation}

Term mutation never replaces a term by a random tree. The term mutation is defined as follows:

\[
if \; X \; is \; Symbolic, \; then \; X \; is \; replaced \; by \; random\; X_i \; i \in [1,n], \; where \; n \; is \; number \; of \; features \; in \; dataset
\]
\[
if \; X \; is \; Constant, \; then \; X \; is \; replaced \; by \; X + r \; where \; r \; is \; normal \; random \; variable \; (0; \; 1) 
\]

\clearpage
\newpage

\subsection{Mutation Probabilities}

During mutation each operator type is being randomly chosen with following probability table:

\bigskip

\begin{tabular}{ |c|c| } 
 \hline
 Operator Type & Probability \\
  \hline
 Boolean & 0.1 \\
   \hline
 Comparison & 0.2 \\
   \hline
 Mathematical & 0.3 \\
    \hline
 Terms & 0.5 \\
   \hline
\end{tabular}

\bigskip
\bigskip

\subsection{Selection}

Rank selection with elite size 1 used as the selection operation \cite{Koza}.
\bigskip
\bigskip

\subsection{Evolution Algorithm}

Evolution is implemented via canonical genetic algorithm method \cite{ga_in_search}:

\bigskip

\begin{algorithm}[H]
\SetAlgoLined
\KwResult{Best individual}
 max\verb|_|generation = 100\;
 population\verb|_|size = 100\;
 cx\verb|_|prob = 0.5\;
 mut\verb|_|prob = 0.5\;
 population = random\verb|_|population(population\verb|_|size)\;
 best\verb|_|ind = select\verb|_|best(population)\;
 generation = 0\;
 \While{best\_ind.fitness < 1 or generation < max\_generation}{
  selected\_population = selection(population) \;
  crossed\_population = crossover(selected\_population, cx\_prob) \;
  mutated\_population = mutation(crossed\_population, mut\_prob) \;
  population = mutated\_population 
  best\_ind = select\_best(population)\; 
  generation++\;
 }
 \Return best\verb|_|ind
 \caption{GP Classifier Evolution Algorithm}
\end{algorithm}

\clearpage
\newpage

\section{Soft Genetic Programming Classifier Design}
\label{sec:sgp_arch}

The main idea of SGP is the usage of weighted continuous functions instead of discontinuous boolean and comparison functions. Weight setting will allow calibrating each operator's effect in a pseudo logical tree, leading to a more adaptive classifier design.

\subsection{Soft Operators}

We change boolean and comparison operators to weighted continuous analogs \cite{o'donnell_2014} (we will call those function sets \textit{pseudo boolean} and \textit{pseudo comparison} operators):

\bigskip

\textbf{Pseudo Boolean}:

\bigskip

 \begin{tikzpicture}[node distance=1mm]
        \node (functionName) at (0, 0) {$OR$:};
        \node[right = of functionName] (domain) {$[0,1]\times[0,1]^2$};
        \node[below = 2mm of domain] (element) {$(w;x,y)$};
        \path (element)--(domain)node[midway,sloped] {$\in$};
        \node[right = 1cm of domain] (codomain) {$[0,1]$};
        \node at (element-|codomain) (image) {$w \cdot max(x,y)$};
        \path (image)--(codomain)node[midway,sloped] {$\in$};
        \draw[->] (domain) -- (codomain);
        \draw[|->] (element) -- (image);
    \end{tikzpicture}

\bigskip

 \begin{tikzpicture}[node distance=1mm]
        \node (functionName) at (0, 0) {$AND$:};
        \node[right = of functionName] (domain) {$[0,1]\times[0,1]^2$};
        \node[below = 2mm of domain] (element) {$(w;x,y)$};
        \path (element)--(domain)node[midway,sloped] {$\in$};
        \node[right = 1cm of domain] (codomain) {$[0,1]$};
        \node at (element-|codomain) (image) {$w \cdot min(x,y)$};
        \path (image)--(codomain)node[midway,sloped] {$\in$};
        \draw[->] (domain) -- (codomain);
        \draw[|->] (element) -- (image);
    \end{tikzpicture}
    
\bigskip

 \begin{tikzpicture}[node distance=1mm]
        \node (functionName) at (0, 0) {$NOT$:};
        \node[right = of functionName] (domain) {$[0,1]\times[0,1]$};
        \node[below = 2mm of domain] (element) {$(w;x)$};
        \path (element)--(domain)node[midway,sloped] {$\in$};
        \node[right = 1cm of domain] (codomain) {$[0,1]$};
        \node at (element-|codomain) (image) {$w(1 - x)$};
        \path (image)--(codomain)node[midway,sloped] {$\in$};
        \draw[->] (domain) -- (codomain);
        \draw[|->] (element) -- (image);
    \end{tikzpicture}
    
\bigskip

Also we will add additional operators:

\bigskip

 \begin{tikzpicture}[node distance=1mm]
        \node (functionName) at (0, 0) {$AND3$:};
        \node[right = of functionName] (domain) {$[0,1]\times[0,1]^3$};
        \node[below = 2mm of domain] (element) {$(w;x,y,z)$};
        \path (element)--(domain)node[midway,sloped] {$\in$};
        \node[right = 1cm of domain] (codomain) {$[0,1]$};
        \node at (element-|codomain) (image) {$w \cdot min(x,y,z)$};
        \path (image)--(codomain)node[midway,sloped] {$\in$};
        \draw[->] (domain) -- (codomain);
        \draw[|->] (element) -- (image);
    \end{tikzpicture}
    
    \bigskip

 \begin{tikzpicture}[node distance=1mm]
        \node (functionName) at (0, 0) {$OR3$:};
        \node[right = of functionName] (domain) {$[0,1]\times[0,1]^3$};
        \node[below = 2mm of domain] (element) {$(w;x,y,z)$};
        \path (element)--(domain)node[midway,sloped] {$\in$};
        \node[right = 1cm of domain] (codomain) {$[0,1]$};
        \node at (element-|codomain) (image) {$w \cdot max(x,y,z)$};
        \path (image)--(codomain)node[midway,sloped] {$\in$};
        \draw[->] (domain) -- (codomain);
        \draw[|->] (element) -- (image);
    \end{tikzpicture}
    
        \bigskip
        
\bigskip

\textbf{Pseudo Comparison}:

\bigskip

 \begin{tikzpicture}[node distance=1mm]
        \node (functionName) at (0, 0) {$>$:};
        \node[right = of functionName] (domain) {$[0,1]\times\mathbb{R}^2$};
        \node[below = 2mm of domain] (element) {$(w;x,y)$};
        \path (element)--(domain)node[midway,sloped] {$\in$};
        \node[right = 1cm of domain] (codomain) {$[0,1]$};
        \node at (element-|codomain) (image) {$w(\frac{x-y}{|x-y|})$};
        \path (image)--(codomain)node[midway,sloped] {$\in$};
        \draw[->] (domain) -- (codomain);
        \draw[|->] (element) -- (image);
    \end{tikzpicture}
    
\bigskip

 \begin{tikzpicture}[node distance=1mm]
        \node (functionName) at (0, 0) {$<$:};
        \node[right = of functionName] (domain) {$[0,1]\times\mathbb{R}^2$};
        \node[below = 2mm of domain] (element) {$(w;x,y)$};
        \path (element)--(domain)node[midway,sloped] {$\in$};
        \node[right = 1cm of domain] (codomain) {$[0,1]$};
        \node at (element-|codomain) (image) {$w(\frac{y-x}{|x-y|})$};
        \path (image)--(codomain)node[midway,sloped] {$\in$};
        \draw[->] (domain) -- (codomain);
        \draw[|->] (element) -- (image);
    \end{tikzpicture}
    
\bigskip

\textbf{Mathematical}:

\bigskip

 \begin{tikzpicture}[node distance=1mm]
        \node (functionName) at (0, 0) {$+$:};
        \node[right = of functionName] (domain) {$\mathbb{R}^2$};
        \node[below = 2mm of domain] (element) {$(x,y)$};
        \path (element)--(domain)node[midway,sloped] {$\in$};
        \node[right = 1cm of domain] (codomain) {$\mathbb{R}$};
        \node at (element-|codomain) (image) {$x + y$};
        \path (image)--(codomain)node[midway,sloped] {$\in$};
        \draw[->] (domain) -- (codomain);
        \draw[|->] (element) -- (image);
    \end{tikzpicture}
    
\bigskip

 \begin{tikzpicture}[node distance=1mm]
        \node (functionName) at (0, 0) {$\Cross$:};
        \node[right = of functionName] (domain) {$\mathbb{R}^2$};
        \node[below = 2mm of domain] (element) {$(x,y)$};
        \path (element)--(domain)node[midway,sloped] {$\in$};
        \node[right = 1cm of domain] (codomain) {$\mathbb{R}$};
        \node at (element-|codomain) (image) {$xy$};
        \path (image)--(codomain)node[midway,sloped] {$\in$};
        \draw[->] (domain) -- (codomain);
        \draw[|->] (element) -- (image);
    \end{tikzpicture}
    
\bigskip

 \begin{tikzpicture}[node distance=1mm]
        \node (functionName) at (0, 0) {$-$:};
        \node[right = of functionName] (domain) {$\mathbb{R}$};
        \node[below = 2mm of domain] (element) {$x$};
        \path (element)--(domain)node[midway,sloped] {$\in$};
        \node[right = 1cm of domain] (codomain) {$\mathbb{R}$};
        \node at (element-|codomain) (image) {$-x$};
        \path (image)--(codomain)node[midway,sloped] {$\in$};
        \draw[->] (domain) -- (codomain);
        \draw[|->] (element) -- (image);
    \end{tikzpicture}

\bigskip

Also we will add additional nonlinear operators:

\bigskip

 \begin{tikzpicture}[node distance=1mm]
        \node (functionName) at (0, 0) {$sigm$:};
        \node[right = of functionName] (domain) {$\mathbb{R}$};
        \node[below = 2mm of domain] (element) {$x$};
        \path (element)--(domain)node[midway,sloped] {$\in$};
        \node[right = 1cm of domain] (codomain) {$\mathbb{R}$};
        \node at (element-|codomain) (image) {$\frac{\mathrm{1} }{\mathrm{1} + e^(-x) }$};
        \path (image)--(codomain)node[midway,sloped] {$\in$};
        \draw[->] (domain) -- (codomain);
        \draw[|->] (element) -- (image);
    \end{tikzpicture}

\bigskip

 \begin{tikzpicture}[node distance=1mm]
        \node (functionName) at (0, 0) {$lin2$:};
        \node[right = of functionName] (domain) {$\mathbb{R}^2\times\mathbb{R}^2$};
        \node[below = 2mm of domain] (element) {$(a,b;x,y)$};
        \path (element)--(domain)node[midway,sloped] {$\in$};
        \node[right = 1cm of domain] (codomain) {$\mathbb{R}$};
        \node at (element-|codomain) (image) {$ax + by$};
        \path (image)--(codomain)node[midway,sloped] {$\in$};
        \draw[->] (domain) -- (codomain);
        \draw[|->] (element) -- (image);
    \end{tikzpicture}
    
\bigskip

 \begin{tikzpicture}[node distance=1mm]
        \node (functionName) at (0, 0) {$lin3$:};
        \node[right = of functionName] (domain) {$\mathbb{R}^3\times\mathbb{R}^3$};
        \node[below = 2mm of domain] (element) {$(a_i;x_i)$};
        \path (element)--(domain)node[midway,sloped] {$\in$};
        \node[right = 1cm of domain] (codomain) {$\mathbb{R}$};
        \node at (element-|codomain) (image) {$\sum{a_ix_i}$};
        \path (image)--(codomain)node[midway,sloped] {$\in$};
        \draw[->] (domain) -- (codomain);
        \draw[|->] (element) -- (image);
    \end{tikzpicture}
    
\bigskip
\textbf{Terms}:
\[
Symbolic:\  \rightarrow X_i \; \; \;   i \in 1,n , \; where \;  n\;  -\;  number\;  of\;  features
\]
\[
Constant:\  \rightarrow \mathbb{R}
\]

\bigskip

\textbf{Weights}:
\[
Weight:\  \rightarrow w \;, \; where \; w \; is \; random \; variable \; uniformly \; distributed \; on \; [0,1]
\]

\clearpage
\newpage

\subsection{Soft Tree Representation}

SGP tree is a tree where each pseudo boolean and comparison operator has its weight parameter, as is shown on Figure \ref{fig:sgp_tree}:

\begin{figure}[H]
\vspace{-2mm}
\centering
    \includegraphics[width = 8cm]{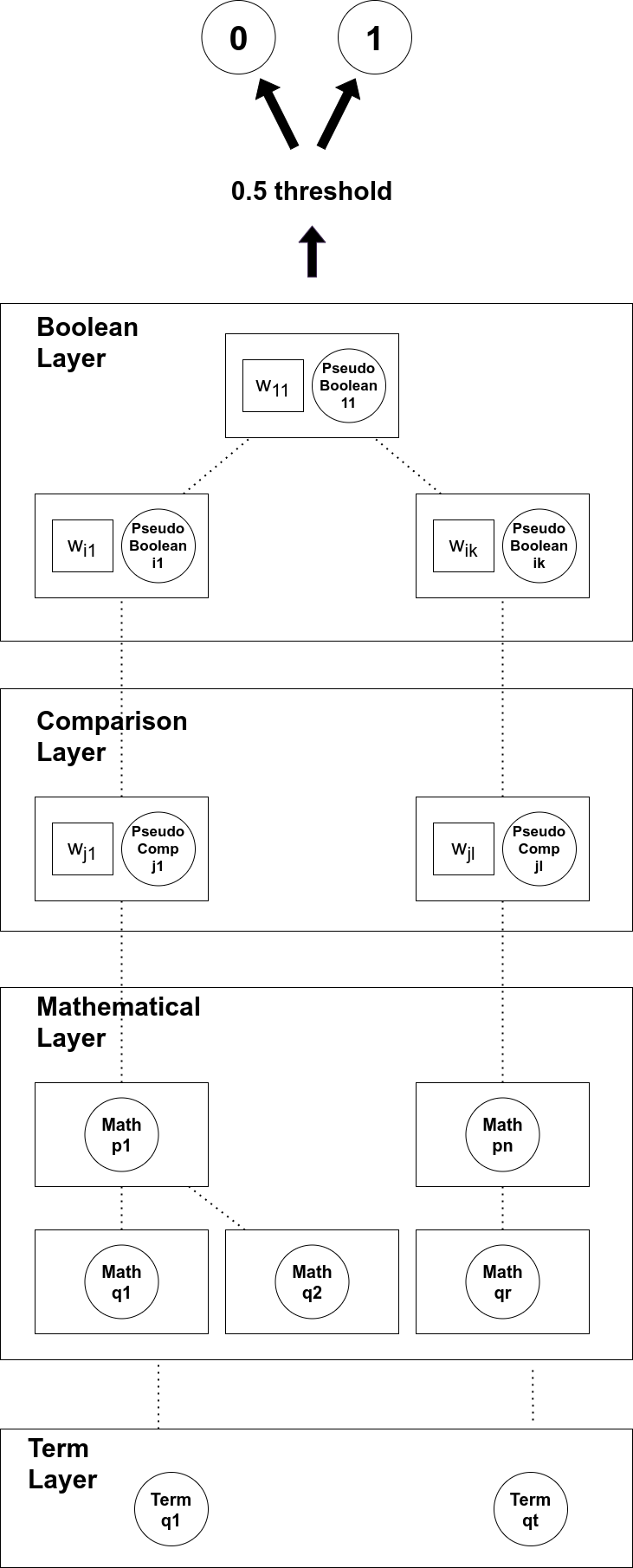}\par 
\caption{SGP Tree}
\label{fig:sgp_tree}
\vspace{-6mm}
\end{figure}

\bigskip

\subsection{Random Tree Generation}

See \hyperref[subsec:gp_random_tree]{3.2}

\subsection{Fitness Function}

See \hyperref[subsec:gp_fitness_function]{3.3}

\subsection{Weight Adjustment}

In SGP we introduce \textit{weight adustment} operation:

\bigskip

\begin{algorithm}[H]
\begin{algorithmic}
\SetAlgoLined
\Procedure{weight\_adjustment}{$individual, max\_tries$}\;
 \For{i in max\_tries}{
  candidate = copy(individual)\;
  coordinate, w = individual.get\_random\_weight()\;
  new\_w = w + random\_shift()\;
  candidate.set\_weight(coordinate, new\_w)\;
  \If{candidate.fitness > individual.fitness} {
     \Return candidate
  }
 }
 \Return individual
 \EndProcedure
 \caption{Weight Adjustment}
\end{algorithmic}
\end{algorithm}

\bigskip

The main point of \textit{weight adjustment} operation is to find any positive improvement using weight calibration.

\bigskip

\subsection{Fitness Driven Genetic Operations}

The main problem of classifier trees is that they have very fragile structures, and the probability of degradation after canonical crossover and mutation operations is to high. In the SGP evolution algorithm, we will use only positive improvement evolution operations. 

\bigskip

\textbf{Positive Crossover} doesn't accept an offsrping which is worse then its parents:

\bigskip

\begin{algorithm}[H]
\begin{algorithmic}
\SetAlgoLined
\Procedure{positive\_crossover}{$ind1, ind2$}\;
  child1, child2 = crossover(ind1, ind2)\;
  candidate1, candidate2 = max([ind1,ind2,child1,child2], by=fitness, 2)\;
 \Return [candidate1, candidate2]
 \EndProcedure
 \caption{Positive Crossover}
\end{algorithmic}
\end{algorithm}

\bigskip

\textbf{Positive Mutation} accepts only those mutations which improves individual's genome: 

\bigskip

\begin{algorithm}[H]
\begin{algorithmic}
\SetAlgoLined
\Procedure{positive\_mutation}{$ind1, max\_tries$}\;
  \For{i in max\_tries} {
    mutant = mutate(ind)\;
    \If{mutant.fitness > ind.fitness} {
		\Return mutant    
    }  
  }
 \Return ind
 \EndProcedure
 \caption{Positive Mutation}
\end{algorithmic}
\end{algorithm}

\clearpage
\newpage

\subsection{Extension Mutation}

\begin{figure}[H]
\vspace{-2mm}
\centering
    \includegraphics[width = 6cm]{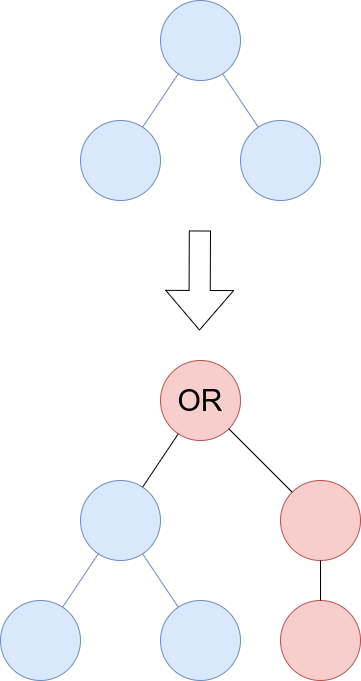}\par 
\caption{Extension Mutation}
\label{fig:ex_mutation}
\vspace{-6mm}
\end{figure}

There is a handy technique for avoiding stucking an improvement of the population in some specific random subspace. A good way of an individual improvement is adding OR operator as tree root with random subtree, as it is shown on the Figure \ref{fig:ex_mutation}.

\bigskip

\bigskip

The \textit{Extension Mutation} operation increases the probability of positive genome improvement.

\clearpage
\newpage

\subsection{Multiple Population}
\bigskip

The usage of fitness driven crossover and mutation operation provokes the problem of lacking gene variation. This problem is solved using the multiple population technique. On each \textit{Nth} generation the best individual of each population is "thrown" to the sequent population \cite{2012.00513}.

\bigskip

\begin{figure}[H]
\vspace{-2mm}
\centering
    \includegraphics[width = 10cm]{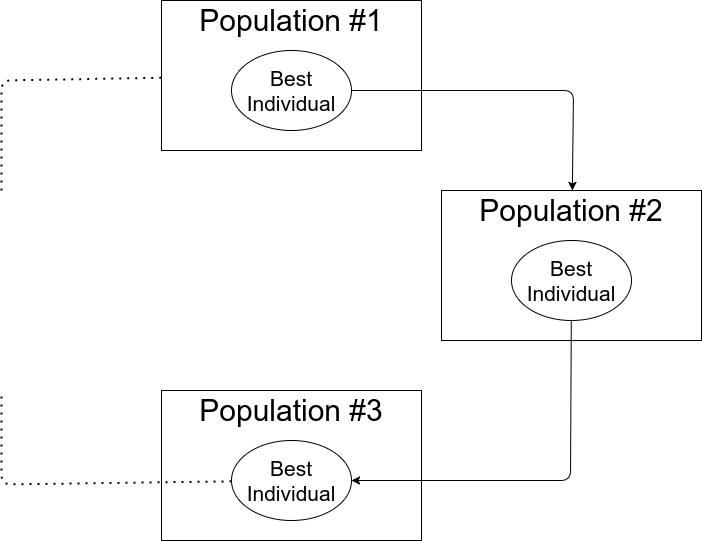}\par 
\caption{Multiple Population}
\label{fig:multiple_population}
\vspace{-6mm}
\end{figure}

\clearpage
\newpage

\subsection{Evolution Algorithm}

SGP evolution algorithm:

\bigskip

\begin{algorithm}[H]
\SetAlgoLined
\KwResult{Best individual}
 max\verb|_|generation = 100\;
 population\verb|_|size = 100\;
 population\_num = 4\;
 cx\verb|_|prob = 0.5\;
 mut\verb|_|prob = 0.5\;
 
 populations = [] \;
 \For{i in population\_num} {
   populations[i] = random\_population(population\_size)\;
 }
 
 best\verb|_|inds = [] \;
 \For{i in population\_num} {
    best\_inds[i] = select\_best(populations[i])\;
 }
 
 best\verb|_|ind\verb|_|ever = max(best\verb|_|inds, key = 'fitness')
 
 generation = 0\;
 \While{best\_ind\_ever.fitness < 1 or generation < max\_generation}{
  \For{i in population\_num} {
  selected\_population = selection(populations[i]) \;
  crossed\_population = crossover(selected\_population, cx\_prob) \;
  mutated\_population = mutation(crossed\_population, mut\_prob) \;
  weighted\_population = weight\_adjustment(mutated\_population) \;
  extended\_population = weight\_adjustment(weighted\_population) \;
  populations[i] = extended\_population \;
  best\_inds[i] = select\_best(populations[i])\; 
  }
  \If{generation mod 5 == 0} {
  \For{i in population\_num} {
    populations[i+1].append(best\_inds[i])\;
 }
 }
  best\_ind\_ever = max(best\_inds, key = 'fitness')\;
  generation++\;
 }
 \Return best\verb|_|ind\verb|_|ever
 \caption{SGP Classifier Evolution Algorithm}
\end{algorithm}

\clearpage
\newpage

\clearpage
\newpage
\section{Visualization}
\label{sec:viz}

Let's compare the behavior of GP and SGP Classifier on generated 2D datasets \cite{sklearn_classifier_comparison}.

\bigskip

\begin{figure}[H]
\vspace{-2mm}
\centering
    \includegraphics[width = 11.5cm]{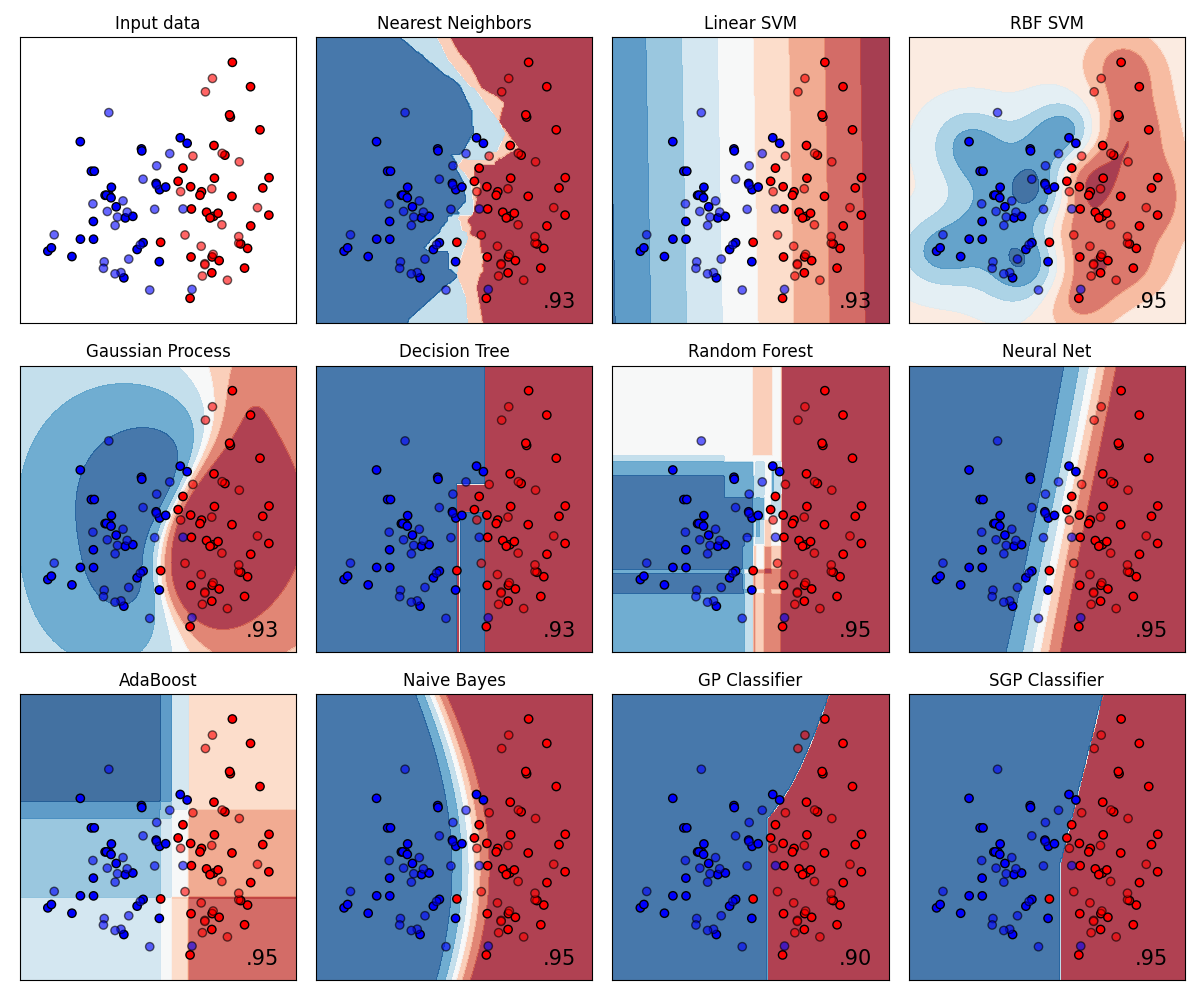}\par 
\caption{Linearly separable dataset}
\label{fig:viz_lin_sep}
\vspace{-6mm}
\end{figure}

\bigskip

\begin{figure}[H]
\vspace{-2mm}
\centering
    \includegraphics[width = 11.5cm]{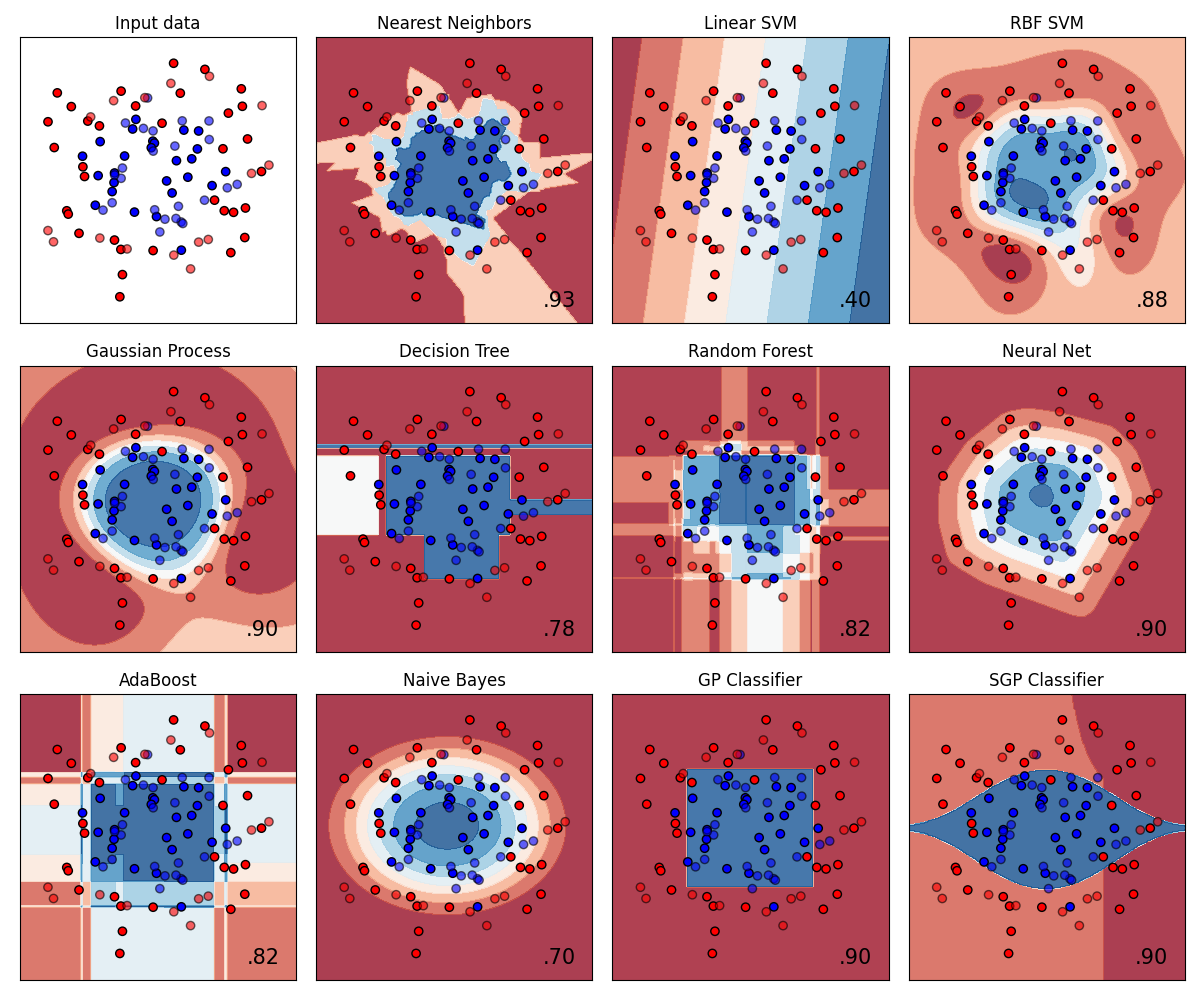}\par 
\caption{Large circle containing a smaller circle dataset}
\label{fig:methodology}
\vspace{-6mm}
\end{figure}

\bigskip

\begin{figure}[H]
\vspace{-2mm}
\centering
    \includegraphics[width = 11.5cm]{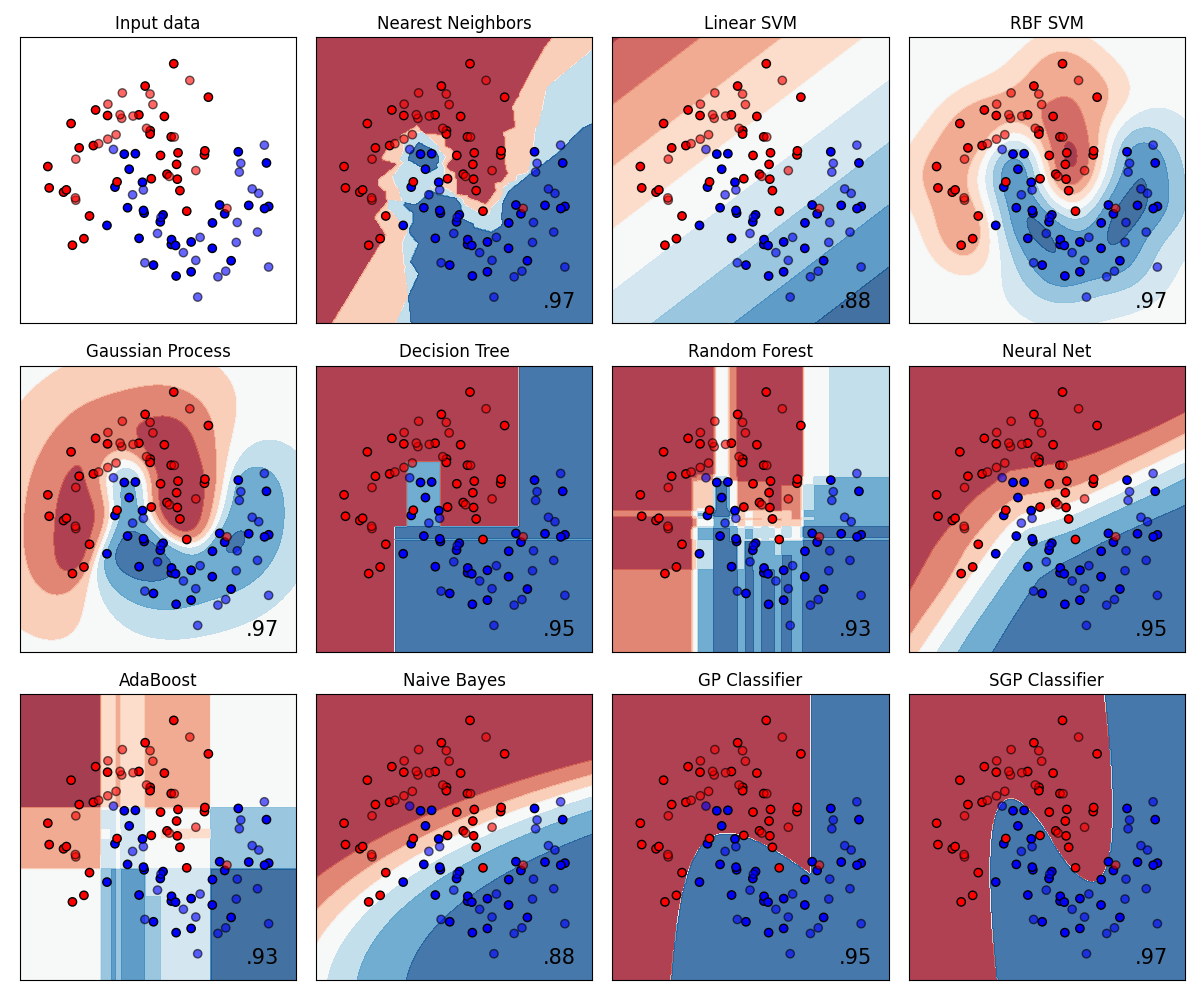}\par 
\caption{Two interleaving half circles dataset}
\label{fig:methodology}
\vspace{-6mm}
\end{figure}

\bigskip

We can see that both GP and SGP classifiers show confident behavior. As it could be expected, the SGP Classifier tends to use nonlinear dependencies. SGP classifier highly likely have strict borders(i.e. $ \mu(\{ SGP(x,y) \in ]0,1[\ \}) \sim  0 $), as it is common behavior for Decision Trees.

\clearpage
\newpage
\section{Experimental Results}
\label{sec:results}

\bigskip

We have tested SGP and GP classifiers using Large Benchmark Suite \cite{1703.00512} for binary classification problem. As a classification quality score we used balanced accuracy.

\bigskip

Test results were gathered by following testing algorithm:

\bigskip

\begin{algorithm}[H]
\SetAlgoLined
 \For{dataset in datasets}{
  \For{i in [1,20]}{
  shuffledDataset = shuffle(dataset) \;
  train, test = split(shuffledDataset, .7) \;
  \For{cls in classifiers}{
    cls.fit(train)
    score = balancedAccuracy(cls, test)
  }
  }
 }
 \caption{Testing algorithm}
\end{algorithm}

\bigskip

\begin{figure}[H]
\vspace{-2mm}
\centering
    \includegraphics[width = \linewidth]{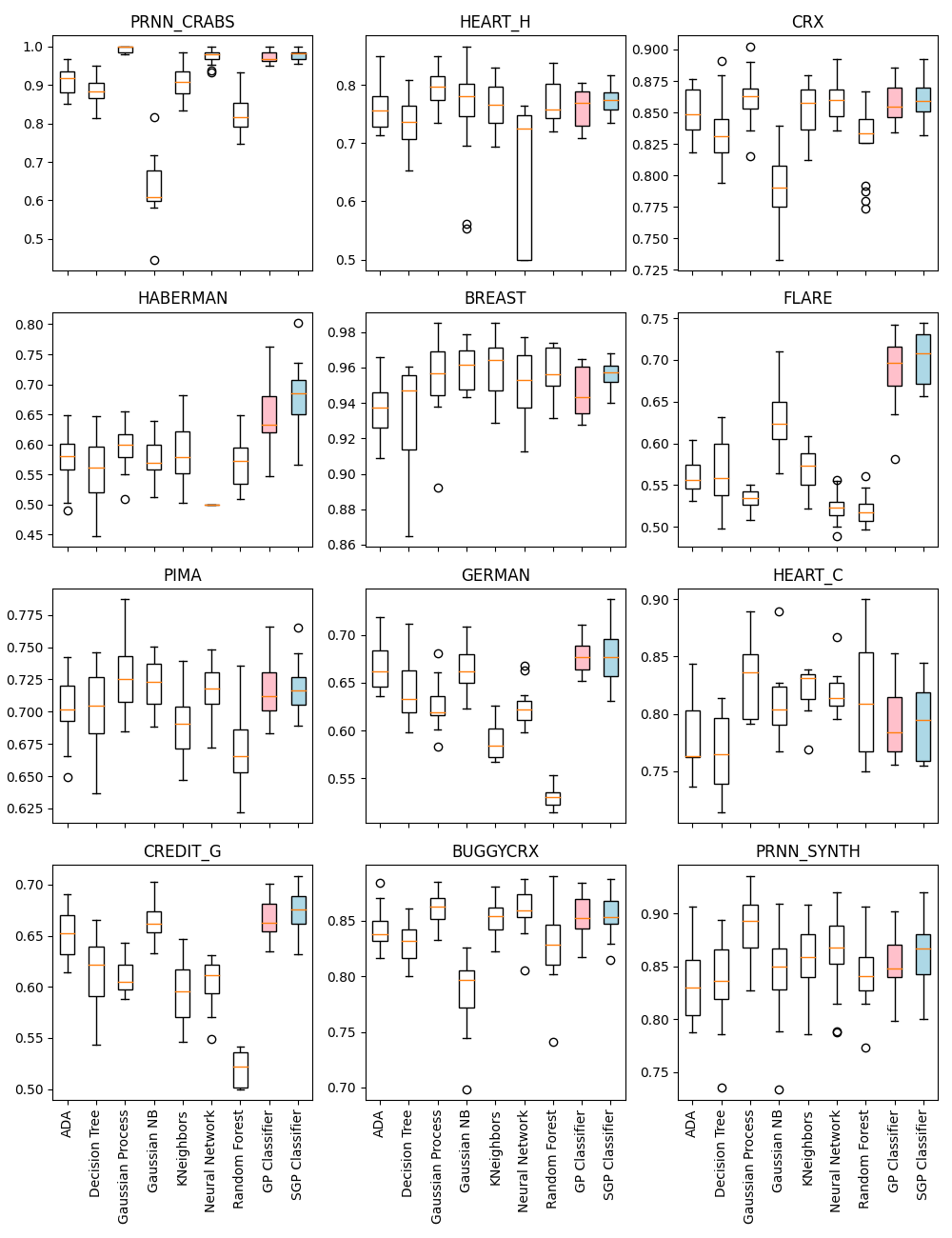}\par 
\caption{Test Results. GP Classifier - pink, SGP Classifer - blue}
\label{fig:results_box}
\vspace{-6mm}
\end{figure}

\clearpage
\newpage

Below we provide a heatmap table view with mean balanced accuracy results:

\bigskip

\def\arraystretch{2}%

\begin{tabular}{ |c|c|c|c|c|c|c| } 
 \hline
  & \textbf{prnn crabs} & \textbf{heart h} & \textbf{crx} & \textbf{haberman} & \textbf{breast} & \textbf{flare} \\ 
  
   \hline   
  \textbf{SGP Classifier} & 
  \cellcolor{red!35}0.978 & 
  \cellcolor{red!35}0.7752 & 
 \cellcolor{red!35} 0.7752 & 
 \cellcolor{red!55}  0.6792& 
  0.9559&
 \cellcolor{red!55}  0.7023   \\ 

  \hline 
  \textbf{GP Classifier} & 
  0.9724 & 
  0.7597& 
  0.7597&
 \cellcolor{red!35}  0.6522& 
       \cellcolor{blue!15}0.9464&   
 \cellcolor{red!35}  0.6856 \\ 

  \hline \textbf{ADA} & 
  0.9095 & 
  0.7594& 
  0.7594& 
  0.5752& 
      \cellcolor{blue!35}0.9358&   
  0.5613\\ 

  \hline \textbf{Decision Tree} & 
  \cellcolor{blue!15}0.8815 & 
  \cellcolor{blue!35}0.7312& 
  \cellcolor{blue!35}  0.7312& 
    \cellcolor{blue!35}0.5585& 
    \cellcolor{blue!55}  0.9328&   
  0.5654\\ 

  \hline \textbf{Gausian Process} & 
  \cellcolor{red!55} 0.994 & 
  \cellcolor{red!55} 0.7925& 
\cellcolor{red!55}   0.7925& 
 \cellcolor{red!15}  0.5981& 
  0.953&   
   \cellcolor{blue!15} 0.534\\ 

  \hline \textbf{Gausian NB} & 
  \cellcolor{blue!55}0.632 & 
  \cellcolor{blue!15}0.748& 
    \cellcolor{blue!15}0.745& 
  0.5748& 
  \cellcolor{red!55}  0.9605&   
 \cellcolor{red!15}  0.6305\\ 

  \hline \textbf{KNeighbors} & 
  0.9079 & 
  0.7654& 
  \cellcolor{red!15}0.7654& 
  0.5857& 
    \cellcolor{red!35}0.96&   
  0.5727\\ 

  \hline \textbf{Neural Network} & 
  \cellcolor{red!15}0.9734 & 
  \cellcolor{blue!55}0.6429& 
  \cellcolor{blue!55}  0.6429& 
  \cellcolor{blue!55}  0.5& 
  0.9517&   
   \cellcolor{blue!35} 0.5229\\ 

  \hline \textbf{Random Forest} & 
  \cellcolor{blue!35}0.8249 & 
  \cellcolor{red!15}0.769& 
  0.769& 
    \cellcolor{blue!15}0.5702& 
     \cellcolor{red!15}0.958&   
  \cellcolor{blue!55}   0.5196\\

 \hline
\end{tabular}

\bigskip

\begin{tabular}{ |c|c|c|c|c|c|c| } 
 \hline
  & \textbf{pima} & \textbf{german} & \textbf{heart c} & \textbf{credit g} & \textbf{buddyCrx} & \textbf{prnn synth} \\ 
  
   \hline   
  \textbf{SGP Classifier} & 
\cellcolor{red!15}0.7181& 
\cellcolor{red!55}0.6791& 
\cellcolor{blue!15}0.7929& 
\cellcolor{red!55}0.674& 
\cellcolor{red!15}0.8559&
\cellcolor{red!35} 0.8642\\ 

  \hline 
  \textbf{GP Classifier} & 
0.7176& 
\cellcolor{red!35}0.6778& 
0.7938&
\cellcolor{red!35}0.6668& 
 0.8537&   
0.8543\\ 

  \hline \textbf{ADA} & 
0.703& 
\cellcolor{red!15}0.6667& 
\cellcolor{blue!35}0.782& 
0.6536& 
0.8427&   
\cellcolor{blue!55}0.8342\\ 

  \hline \textbf{Decision Tree} & 
\cellcolor{blue!15}0.7009& 
0.6394& 
\cellcolor{blue!55}0.7661& 
0.6151& 
\cellcolor{blue!15}0.8309&   
\cellcolor{blue!35}0.8368\\ 

  \hline \textbf{Gausian Process} & 
\cellcolor{red!55}0.7288& 
\cellcolor{blue!15}0.6255& 
\cellcolor{red!55}0.8304& 
0.6101& 
\cellcolor{red!55}0.8614&   
\cellcolor{red!55}0.8886\\ 

  \hline \textbf{Gausian NB} & 
\cellcolor{red!35}0.7218& 
0.6633& 
0.8127& 
\cellcolor{red!15}0.6637& 
\cellcolor{blue!55}0.7866&   
\cellcolor{blue!15}0.844\\ 

  \hline \textbf{KNeighbors} & 
\cellcolor{blue!35}0.6886& 
\cellcolor{blue!35}0.588& 
\cellcolor{red!15} 0.8191& 
\cellcolor{blue!35}0.5941& 
0.8517&   
0.8551\\ 

  \hline \textbf{Neural Network} & 
0.7168& 
\cellcolor{blue!15}0.6255& 
\cellcolor{red!35}0.8206& 
\cellcolor{blue!15}0.6041& 
\cellcolor{red!35}0.8598&   
\cellcolor{red!15}0.8629\\ 

  \hline \textbf{Random Forest} & 
\cellcolor{blue!55}0.6714& 
\cellcolor{blue!55}0.5297& 
0.8146& 
\cellcolor{blue!55}0.5197& 
\cellcolor{blue!35}0.828&   
 0.846\\

 \hline
\end{tabular}

\bigskip
\bigskip

SGP provides very positive average results, and especially nice at \textit{haberman} and \textit{flare} datasets. All results can be regathered running script \url{https://github.com/survexman/sgp_classifier/blob/main/soft/gp_classification.py}

\clearpage
\newpage
\section{Conclusion}

\bigskip

This survey introduces the concept of Soft Genetic Programming. The robustness of the SGP Classifier is presented. This research aimed to deliver a new genetic programming design with pseudo boolean and pseudo comparison operators and show its quality.

\bigskip
\bigskip

Never the less SGP has several drawbacks:
\begin{itemize}
  \item Performance issue. SGP training stage takes much more time than classical classifers do.
  \item  An incapability to use SGP as a feature selection tool. Due to the weighted operator design, each branch and operator can have a very low weight. Thus the term that belongs to the weighted operator has low significance and cannot be used as a significant feature. In classical GP Classifier, each symbolic variable highly likely has a high significance degree and can thus be selected as meaningful feature.
\end{itemize}

\bigskip

Anyway, the improvement of the SGP technique for a particular task can provide excellent practical results.

\end{document}